# AN APPROACH TO IMPROVING EDGE DETECTION FOR FACIAL AND REMOTELY SENSED IMAGES USING VECTOR ORDER STATISTICS

B O. Sadiq, S.M. Sani and S. Garba

Department of Electrical and Computer Engineering,
Ahmadu Bello University, Zaria

*ABSTRACT*

*This paper presents an improved edge detection algorithm for facial and remotely sensed images using vector order statistics. The developed algorithm processes coloured images directly without been converted to grey scale. A number of the existing algorithms converts the coloured images into grey scale before detection of edges. But this process leads to inaccurate precision of recognized edges, thus producing false and broken edges in the output edge map. Facial and remotely sensed images consist of curved edge lines which have to be detected continuously to prevent broken edges. In order to deal with this, a collection of pixel approach is introduced with a view to minimizing the false and broken edges that exists in the generated output edge map of facial and remotely sensed images.*

## 1. INTRODUCTION

One of the most important task in image processing is detection of edges [1]. Edge detection is a low level feature in image processing that deals with the extraction of important features in images. An edge in an image is caused by local discontinuity in pixel due to either light, shadows or illumination [2]. The fundamental goal of edge detection is to produce a line drawing of a scene from an image of that Scene. Thus, important features such as curves and corners can be extracted from the edges of the images[3]. Face detection is a technique used to find faces at different locations with different sizes in a given location. It is applicable in the field of image processing in biometrics, multimedia applications, video surveillance amongst others [4]. The fundamental aim of face and object detection is successful edge identification and extraction. Edge maps are generated in face detection with a view to representing faces as a single unit. These generated edge maps are one of the most popular way of representing facial images and it features [5]. Remotely sensed images are data that contain important information which are acquired about an object or phenomenon without making physical contact. This replaced expensive and inefficient data collection on ground, assuring that areas of process are not disturbed [6]. The Edge maps generated using edge detection algorithms concentrates on the pertinent information of a remotely sensed image, the method used to extract these edge maps effectively is extremely important for image processing applications [7]. Some of the pertinent information generated by applying the edge detection algorithms on remotely sensed images are road networks, geological features, and desert extraction amongst others [8].

Numerous researchers have developed edge detection algorithms for facial and remotely sensed images such as the work of [4], [5], [6], [9] and [10]. The authors in [4] and [5] presented an algorithm for extraction of edges in facial images using the Sobel and Canny edge detection algorithms. But broken and false edges exist in the output edge map. The authors in [6] and [9] also presented an algorithm for edge detection in satellite images. However the algorithm produced displaced edges. The author in [10] presented a comprehensive edge detection algorithm for satellite images using the laplacian mask. This method produced falsified edge lines. In view of the imperfection associated with the existing works, there is need to develop an improved edge detection algorithm that will produce thin and continuous edge lines in the generated output edge maps.

## 2. VECTOR ORDER STATISTICS

A typical way to represent coloured images in a vector form. Coloured images are 3-D images that assign three numerical values to each pixel in an image [11]. The ordering of these numerical channels is defined as Vector Order Statistics. This ordering of component in coloured images are of four different types namely: the Vector Range (VR), Minimum Vector Range (MVR), Vector Dispersion (VD) and Mean Vector Dispersion (MVD) [12]. The easiest to implement and less sensitive to noise is the minimum vector range. The minimum vector range calculates the Euclidean distance between two pixels in an image after ordering of the sort using equation (2.1) [13].

$$MVR = \| X_n - X_1 \| \qquad (2.1)$$

Where; ∥ ∥ is the vector norm,

$X_n$ is the n$^{th}$ pixel in the image

$X_1$ is the last pixel in the image

Figure 2.1 shows the 3x3 window indexing

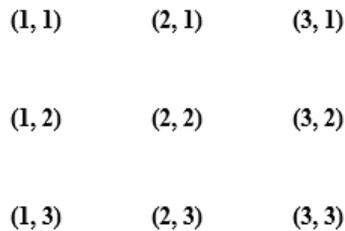

Figure 2.1: Edge Pixel Indexing/ Arrangement

## 3. METHODOLOGY

i. From the input image, generate a 3x3 window size pixel.

ii. Find the Euclidean distance between each pixel in the window

iii. Apply the developed mask based on collection of pixel.

iv. Use non maximum suppression to reduce thick edges.

v. Determine which pixel is an edge or not using a threshold value.

### 3.1 Pixel Collection

With a view to reducing false and broken edges at curves, a collection of pixel scheme is proposed based on the step and roof edge profile as depicted in figure 3.1

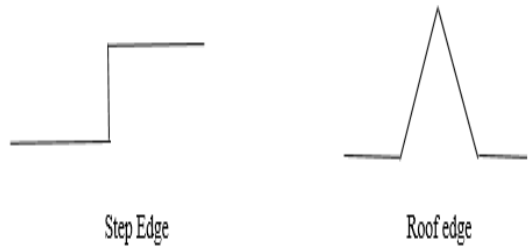

Figure 3.1 Step and Roof Edge Profile

The collection of pixels for each collection scheme are from integers 0-8 which implies that the collection of pixels are that of an 8- Neighbourhood pixel. Figure 3.2 shows the integer notation for 8-Neighborhood Pixel in a 3x3 Window.

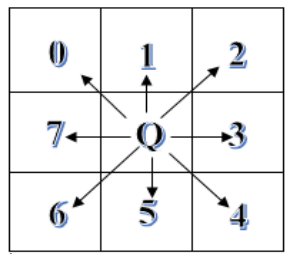

Figure 3.2 Integer Notation for 8-Neighborhood Pixel in a 3x3 Window.

Based on the Step and Roof Profile, a Collection scheme is generated as in Figure 3.3

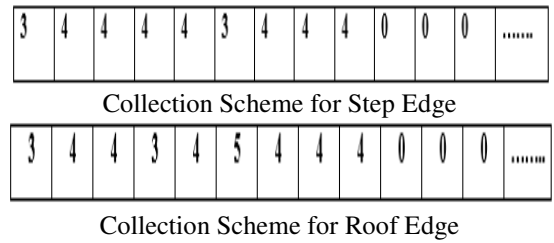

Collection Scheme for Step Edge

Collection Scheme for Roof Edge

Figure 3.3 Collection Scheme based on Step and Roof Edge

The developed collection scheme are represented as a mask and applied to the image with a view to producing thin and continuous edge lines. The generated mask developed from the collection scheme as shown in figure 3.4

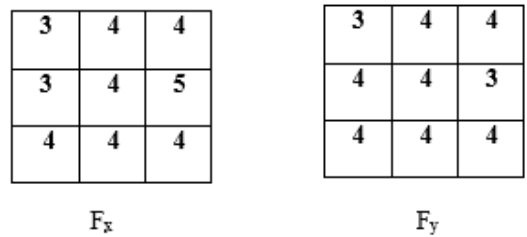

Figure 3.4 Developed Mask from the Collection Scheme

## 3.2 Vector Range

Let R, G, B denote the unit vectors along the RGB axis of the RGB colour space. Given an image I, the vector of the colour space in the image can be defined as

$$m = \frac{\partial R}{\partial P} r + \frac{\partial G}{\partial P} g + \frac{\partial B}{\partial P} b \qquad (3.1)$$

$$n = \frac{\partial R}{\partial Q} r + \frac{\partial G}{\partial Q} g + \frac{\partial B}{\partial Q} b \qquad (3.2)$$

(m, n) = size (I) where, size (I) is the dimension of the image used.

The Euclidean distance of a pixel m is given by

$$\|m\| = \sqrt{m \bullet m} \qquad (3.4)$$

Where, m is in the norm form

In a grid of pixels that constitutes the image, the Euclidean distance between two pixels is (m, n) is

$$D_E(m, n) = \|n - m\| \qquad (3.5)$$

Since the input images is a colour image consisting of three components, this can be represented as

$$D_E(m_1, n_1) = \sqrt{(n_{11} - m_{21})^2 + (n_{12} - m_{22})^2 + (n_{13} - m_{23})^2}$$

Where $D_E$ is the Euclidean distance between two pixels

$$D_E(P_{1,1}RGB - P_{1,2}RGB) = \sqrt{(P_{1,1}R - P_{1,2}R)^2 + (P_{1,1}G - P_{1,2}G)^2 + (P_{1,1}B - P_{1,2}B)^2}$$

The Vector Range between grids of pixels is calculated by subtracting the last pixel for the first pixel

$$VR_1 = |P_{3,2}RGB - P_{1,1}RGB|$$

$$VR_2 = |P_{2,3}RGB - P_{1,1}RGB|$$

$$VR_3 = |P_{1,3}RGB - P_{1,1}RGB|$$

$$VR_4 = |P_{3,1}RGB - P_{1,1}RGB|$$

$$VR_5 = |P_{2,1}RGB - P_{1,1}RGB|$$

$$VR_6 = |P_{1,2}RGB - P_{1,1}RGB|$$

$$VR_7 = |P_{3,3}RGB - P_{1,1}RGB|$$

$$VR_8 = |P_{2,2}RGB - P_{1,1}RGB|$$

$$VR_0 = |P_{1,1}RGB - P_{1,1}RGB|$$

MVR = min {VR0, VR1, VR2, VR3, VR4, VR5, VR6, VR7, VR8}

Where, MVR = Minimum Vector Range

The Flow Chart for the Developed Algorithm is shown in Figure 3.5

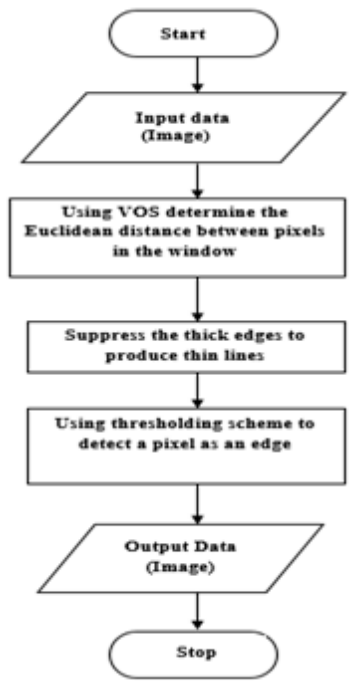

Figure 3.5: Flow Chart for the Developed Algorithm

The sample of the facial and remotely sensed images collected are depicted in figures 3.6 and 3.7

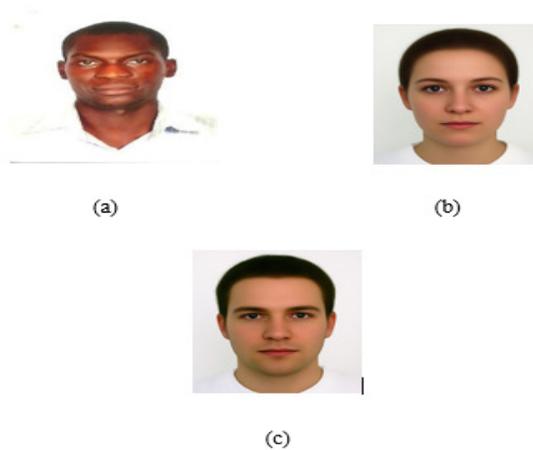

Figure 3.6: Sample of Faces Used

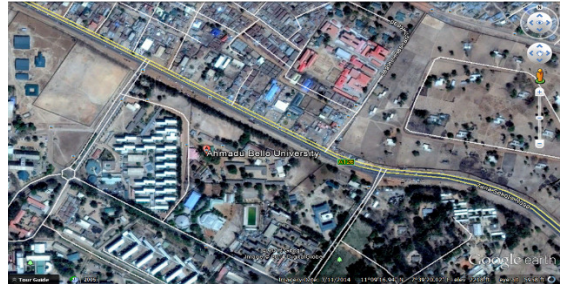

Figure 3.7: Sample of Remotely Sensed Image Used

## 4. RESULTS AND DISCUSSIONS

After applying the developed edge detection algorithm to the sample of images collected, the output result are presented in Figures 4.1 and 4.2

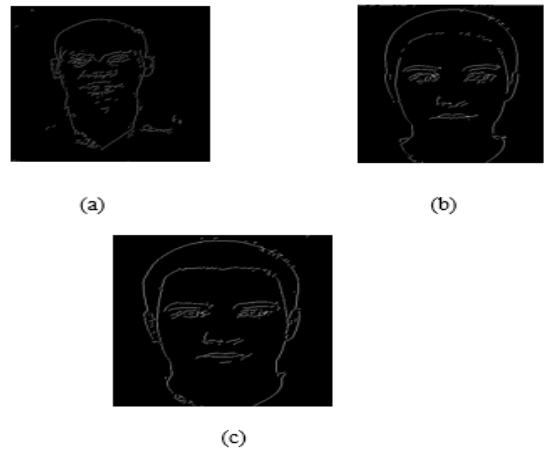

Figure 4.1: Output Result of Applying the proposed Algorithm of Facial Images

Figure 4.1 shows the edge maps of applying the proposed algorithm based on vector order statistics to the sample of collected natural faces used. The visual performance of the result of applying the algorithm showed that the edges that constitute the overall face was extracted completely. The algorithm is not dependent on the type of skin colour in the image. With high resolution images the processing time increases, unless if used on a dedicated and specialized systems. But different facial expression produces different generated edge map for the same individual.

The output result of applying the developed algorithm to remotely sensed image is shown in figure 4.2

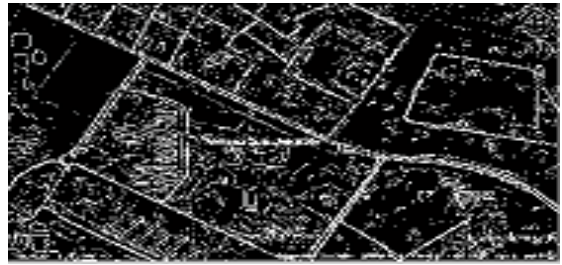

Figure 4.2 Output Result of Applying the proposed Algorithm of Remotely Sensed Images

Figure 4.2 shows the edge maps of applying the proposed algorithm based on vector order statistics to the sample of remotely sensed image used. The visual performance of the result of applying the algorithm showed extracted regions and roads in the remotely sensed image.

## 5. CONCLUSION

This paper presents an approach to improving edge detection in facial and remotely sensed images using a pixel collection scheme. This developed pixel collection scheme is developed with a view to addressing the problem of false and broken edges that exist in these images due to curves. The developed algorithm can be applied on more facial and remotely sensed images to test its effectiveness and compare with other developed existing techniques.